\documentclass[balance,upint,subscriptcorrection,varvw,mathalfa=cal=boondoxo,spanish,french,vietnamese,russian,greek,pdf-a,colorlinks]{asmeconf}

\usepackage{multirow}
\usepackage{multicol}
\usepackage{gensymb}
\usepackage{tabularx}
\usepackage{booktabs,multirow,siunitx}
\hypersetup{%
	pdfauthor={John H. Lienhard},									  
	pdftitle={ASME Conference Paper LaTeX Template},                  
	pdfkeywords={ASME conference paper, LaTeX template, BibTeX style},
	pdfsubject = {Describes the asmeconf LaTeX template},			  
	pdflicenseurl={https://ctan.org/pkg/asmeconf},
}

\begin{document}




\title{Vision–Language Models for Infrared Industrial Sensing in Additive Manufacturing Scene Description}

\SetAuthors{%
	Nazanin Mahjourian\affil{}\CorrespondingAuthor{mahjouri@mtu.edu}, 
	Vinh Nguyen
}

\SetAffiliation{}{Department of Mechanical and Aerospace Engineering,\\
Michigan Technological University, Houghton, MI 49931}




\maketitle



\keywords{Computer Vision, Thermal infrared imaging, Vision-Language Models, Zero Shot Learning, Prompt engineering, Smart Manufacturing, Additive manufacturing monitoring}


\begin{abstract}

Many manufacturing environments operate in low-light conditions or within enclosed machines where conventional vision systems struggle. Infrared cameras provide complementary advantages in such environments. Simultaneously, supervised AI systems require large labeled datasets, which makes zero-shot learning frameworks more practical for applications including infrared cameras. Recent advances in vision–language foundation models (VLMs) offer a new path in zero-shot predictions from paired image–text representations. However, current VLMs cannot understand infrared camera data since they are trained on RGB data. This work introduces VLM-IRIS (Vision–Language Models for InfraRed Industrial Sensing), a zero-shot framework that adapts VLMs to infrared data by preprocessing infrared images captured by a FLIR Boson sensor into RGB-compatible inputs suitable for CLIP-based encoders. We demonstrate zero-shot workpiece presence detection on a 3D printer bed where temperature differences between the build plate and workpieces make the task well-suited for thermal imaging. VLM-IRIS converts the infrared images to magma representation and applies centroid prompt ensembling with a CLIP ViT-B/32 encoder to achieve high accuracy on infrared images without any model retraining. These findings demonstrate that the proposed improvements to VLMs can be effectively extended to thermal applications for label-free monitoring.

\end{abstract}





\begin{figure*}[htb!]
    \centering
    \includegraphics[width=0.8\linewidth, keepaspectratio]{figures/Dataconversion5.pdf}
    \caption{Overview of the problem statement that shows the comparison of RGB and infrared imaging. In dark conditions the printed part is barely visible in RGB images, while the magma representation of the infrared image consistently captures the part.}
    \label{fig:dataconversion}
\end{figure*}

\section{Introduction}

Artificial intelligence (AI) is increasingly transforming manufacturing, where computer vision technologies play central roles in many tasks such as monitoring processes, detecting defects, and ensuring quality~\cite{mozaffar2025tunes,yazdipaz2025robust,safavigerdini2025automated, dashti2025functional,lalehdashti2025constructing}. However, most vision-based systems today depend on large labeled datasets that are built for specific tasks and environments. This reliance makes them hard to adapt in dynamic manufacturing settings, since collecting and labeling data is both time-consuming and labor-intensive\cite{ziad2025advancing}. In addition, many industrial environments are low light or enclosed, which reduces the reliability of standard optical cameras. In enclosed additive manufacturing systems, printed parts can be difficult to distinguish from the build surface when using RGB cameras, especially under poor lighting conditions. Thermal infrared (IR) imaging offers a strong alternative because it does not depend on ambient light and can capture useful contrast from temperature differences between machine surfaces, tools, and parts. Figure~\ref{fig:dataconversion} illustrates this contrast by comparing RGB and infrared images under dark conditions.

Thermal imaging has been widely applied in manufacturing through supervised deep learning approaches~\cite{altaf2022usage}. UIR-Net~\cite{guo2021uir} introduces a pipeline for object detection in thermal IR images that automatically generates labels from unlabeled data using SIFT and clustering before training a deep model. Deep learning predictions can be combined with engineered IR emissivity features to improve classification under varying lighting conditions and materials that appear similar in visible~\cite{grossmann2022improving}. IR signatures can effectively distinguish between normal and faulty states, when the infrared cameras have been paired with detection to monitor manufacturing processes~\cite{wang2023manufacturing}. Optical imaging and infrared thermography can be integrated for monitoring of FFF 3D printing. Thermal hotspots correlate with embedded defects and can enable real-time fault detection~\cite{abouelnour2023assisted}. These methods are very effective in manufacturing, but they rely on domain-specific pipelines and retraining models for each task and also require a large amount of labeled data.

Zero-shot learning (ZSL) offers a promising alternative that enables models to generalize to unseen tasks without the need for retraining~\cite{xian2017zero,aghdam2025actalign}. A ZSL model can make predictions on previously unseen scenarios by using additional information such as descriptive attributes or semantic text descriptions. This approach is especially useful in manufacturing, where new parts, materials, or process conditions are introduced frequently and collecting labeled data for each scenario is often time-consuming and impractical. Hua et al.~\cite{hua2022zero} introduced a framework for tool wear prediction under non-stationary machining environments. They combined deep learning with causal inference to extract invariant features across conditions. A tensor-based method for zero-shot fault diagnosis in smart process manufacturing was also developed to evolve dual prototypes to align sample and attribute representations~\cite{ren2024zero}. They improved generalization to unseen faults across multiple datasets. Huang et al.~\cite{huang2023simple} presented a simple framework for generalized zero-shot fault diagnosis, using binary classifiers with semantic correction to reduce bias in fault attributes. These studies highlight the strength of ZSL in manufacturing because collecting and labeling large datasets is both costly and often impractical.

Recent advances in vision–language models (VLMs) such as CLIP~\cite{radford2021learning}, OpenCLIP~\cite{cherti2023reproducible}, SigLIP~\cite{zhai2023sigmoid}, and EVA-CLIP~\cite{sun2023eva} have taken zero-shot learning to a new level. These models excel in two key areas: (1) they have a broad and diverse knowledge of visual and linguistic concepts because they are pretrained on huge image–text datasets drawn from the web, and (2) they are able to align visual features with natural language descriptions by learning a shared multimodal embedding space. VLMs typically adopt a dual-encoder design with separate encoders for images and text, often implemented using transformers~\cite{wolf2020transformers,ahmadi2025unsupervised}. The image encoder extracts high-level visual representations from input images, while the text encoder processes tokenized sentences and captures their semantic meaning~\cite{kiashemshaki2025simulating,anilkumar2025comparative}. The resulting embeddings are projected into a common latent space, and their similarity can be directly measured. This shared embedding space makes VLMs highly flexible for downstream tasks without requiring task-specific retraining.

Early zero-shot learning methods typically relied on handcrafted attributes such as shape, color, or texture, or on semantic labels that describe class properties to bridge seen and unseen classes. In contrast, VLMs achieve zero-shot transfer by directly matching images with natural-language descriptions. This capability makes VLMs a powerful tool for manufacturing applications. They can be personalized with hybrid prompts for zero-shot anomaly detection by combining textual anomaly descriptions with symbolic rules and region constraints~\cite{cao2025personalizing}. This method achieved state-of-the-art performance on several industrial anomaly datasets and a real-world automotive inspection task. Cai et al.~\cite{caitowards} used detailed anomaly descriptions from multimodal language models together with attention prompts from vision foundation models to develop a hybrid explainable prompt enhancement framework for zero-shot anomaly detection. These methods show that VLMs can support data-efficient quality control in manufacturing when they are guided by tailored prompting strategies~\cite{torkamani2024assertify}. Some studies have shown that strategies such as multiview queries and chain-of-thought reasoning improve recognition across machining, additive manufacturing, and sheet metal domains. Khan et al.~\cite{khan2025leveraging} tested GPT-4o, Claude-3.5, and other VLMs for automatic feature recognition in CAD models to demonstrate this. Another framework achieved state-of-the-art performance by using CLIP for anomaly localization and SAM for precise segmentation in a zero-shot anomaly segmentation task~\cite{li2025clipsam}.

As discussed above, VLMs have recently shown promise in manufacturing, but their use in this domain is still at an early stage~\cite{zhao2025industrial}. A key limitation for utilizing these models in manufacturing tasks is that they are trained almost entirely on natural RGB images~\cite{faysal2024nmformer}. This makes it unclear how well they can transfer to other modalities such as thermal infrared imagery~\cite{borazjani2025multi}. This is an important gap because thermal cameras are already common in low-light or enclosed industrial environments where RGB cameras are unreliable. However, most current industrial implementations rely on supervised pipelines that are customized to individual machines and processes, and large available infrared datasets with semantic labels or descriptive text are still rare. As a result, many existing approaches require custom data collection and retraining for each new task, which limits how easily they can be scaled or adapted. This gap motivates the use of zero-shot and label-free approaches that can build on existing pretrained models without the need for extensive infrared training data. Some studies in other domains have adapted vision foundation models to infrared data, but no prior work has systematically investigated zero-shot classification with VLMs on thermal IR data in manufacturing tasks. 

\begin{figure*}[htb!]
    \centering
    \includegraphics[width=\linewidth, keepaspectratio]{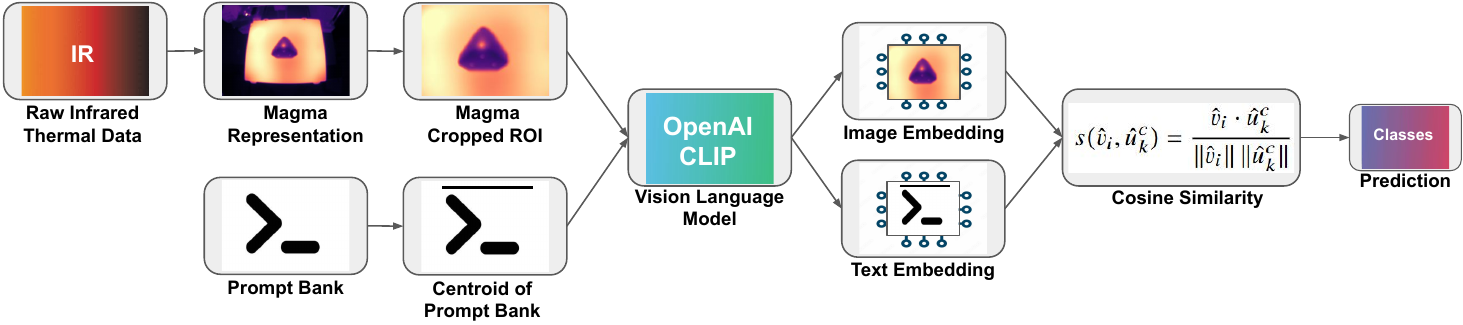}
    \caption{Overview of the VLM-IRIS framework. Raw thermal infrared images are preprocessed into magma representations and cropped to the region of interest. A prompt bank is constructed for the two classes (\textit{present}, \textit{absent}), and prompt embeddings are averaged to form class centroids. Both images and prompts are embedded into a shared multimodal space using CLIP. Cosine similarity between the image embedding and class centroids determines the predicted label.}
    \label{fig:method}
\end{figure*}

Reliable object presence detection on the build platform is an important component of autonomous manufacturing workflows. In additive manufacturing environments with downstream automation, such as robotic pick-and-place or transfer to subsequent processing stages, confirming that a printed part is present on the build plate is a necessary prerequisite for safe and reliable operation. Establishing this capability using thermal infrared sensing and zero-shot vision–language models provides a flexible foundation for automation without requiring task-specific retraining. This work introduces VLM-IRIS (Vision–Language Models for InfraRed Industrial Sensing), a zero-shot framework for adapting VLMs to thermal infrared imagery. It combines the strengths of zero-shot VLMs and infrared imaging to enable reliable object presence detection without any model retraining. The framework converts the infrared images into RGB representations suitable for VLMs to bridge the modality gap. It also utilizes a centroid-based prompting strategy, which multiple descriptive sentences are averaged to create stable text embeddings that reduce sensitivity to prompt wording in order to capture the highest similarity between image-text pairs in the VLMs' semantic shared space. The goal of this work is to better understand how vision–language models can be used with thermal infrared data in manufacturing settings. Specifically, we study whether pretrained vision–language models can be applied to infrared images without retraining, how different infrared-to-RGB preprocessing choices influence zero-shot performance, and how prompt design strategies affect the robustness of infrared scene classification. In particular, we compare preprocessing approaches that preserve the original thermal structure with those that introduce color variations commonly used in scientific visualization, to examine how these choices interact with vision–language models trained on natural RGB imagery. Through these investigations, this work evaluates both the potential and the current limitations of using vision–language models for label-free infrared monitoring in additive manufacturing. The best-performing configuration of the framework combines CLIP with magma preprocessing and centroid-based prompting to achieve strong accuracy in zero-shot classification. To further validate the robustness of VLM-IRIS, we also evaluate alternative settings across different preprocessing pipelines, prompting strategies, and dataset conditions.

The rest of the paper is organized as follows. Section~\ref{sec:method} presents the methodology of the VLM-IRIS framework. Section~\ref{sec:experiments} describes experiments that compare this approach to different preprocessing and prompting strategies. Section~\ref{sec:results} reports experimental results, and Section~\ref{sec:conclusions} provides conclusions and directions for future work.


\section{Method}
\label{sec:method}

The objective is to perform zero-shot classification of thermal infrared images into two categories: \textit{object present} and \textit{object absent}. 
The dataset consists of $I$ infrared images, which is denoted as $\mathcal{X} = \{x_i\}_{i=1}^I$, with the corresponding ground-truth labels $\mathcal{Y} = \{y_i\}_{i=1}^I$, where each $y_i \in \{\text{present}, \text{absent}\}$. The goal is to predict the label $\hat{y}_i$ for each image $x_i$ without training or fine-tuning on infrared data. Instead, the framework adapts pretrained VLMs to RGB-compatible modality through preprocessing and text-image similarity scoring. Figure~\ref{fig:method} illustrates the method proposed in this paper, which is described in detail in the following subsections.

\subsection{Preprocessing and Modality Adaptation}

VLMs such as CLIP are trained on massive datasets of natural RGB images that are paired with text descriptions. Non-RGB modalities, such as raw thermal infrared data that appear as single-channel intensity maps, cannot be directly passed to these models. To make thermal images compatible with these models, each thermal image $x \in \mathcal{X}$ is converted into an RGB representation $\tilde{x}$ using one of three preprocessing approaches: grayscale mapping, magma colormap, and viridis colormap. The preprocessing function is defined as:

\begin{equation}
\tilde{x} = \mathcal{P}_m(x), \quad m \in \{\text{grayscale}, \text{magma}, \text{viridis}\}
\end{equation}

where $x$ is the raw thermal image, $\mathcal{P}_m(\cdot)$ is the mapping function, and $\tilde{x}$ is the resulting three-channel RGB image. This step allows the production of inputs that VLMs can interpret and work with thermal infrared imagery without any retraining.

\subsection{Prompt Banks}

Every task performed with VLMs is expressed as prompts since these models link images with natural-language descriptions. Unlike supervised learning, where labels are learned directly from annotated data, zero-shot inference with VLMs depends entirely on how the categories are described to the model as a prompt. Hence, the categories of interest must be a formulated sentence that describes the scene for the model.

To reduce sensitivity to the exact wording of a single sentence, a prompt bank was formed to describe each class with a set of different sentences. In this bank for each category (\textit{present}, \textit{absent}), multiple descriptive sentences were written to capture variations in phrasing and perspective. For each class $c \in \mathcal{Y}$, the prompt bank is defined as $\{t^c_1, \dots, t^c_K\}$. Using a bank instead of a single description provides broader coverage of linguistic variations and prevents the model from being limited to suboptimal phrasing.

\subsection{Embeddings and Classification via Cosine Similarity}

After preprocessing the infrared images and defining the prompt banks, a pretrained VLM was utilized to project both the RGB images and the prompts into a shared multimodal embedding space. The model employs two separate encoders: one for images and one for text. The image encoder transforms each infrared image into a feature vector while the text encoder maps each prompt into the same space. All vectors were normalized to unit length for evaluation based on their directions. 

Given an image $\tilde{x}_i$, the image encoder produces a feature representation:
\begin{equation}
v_i = f_\theta(\tilde{x}_i), \quad \hat{v}_i = \frac{v_i}{\|v_i\|_2}
\end{equation}
where $f_\theta$ denotes the image encoder and $\hat{v}_i$ is the $L_2$-normalized embedding. 

Similarly, each prompt $t^c_k$ that belongs to class $c$ is mapped into the same embedding space by the text encoder:

\begin{equation}
u^c_k = g_\phi(t^c_k), \quad \hat{u}^c_k = \frac{u^c_k}{\|u^c_k\|_2}  
\end{equation}

Once both images and prompts are embedded in the shared multimodal space, the cosine similarity score is used to measure their alignment. This score measures the closeness of an image and  a text description and emphasizes the orientation of the vectors on the unit hypersphere rather than their magnitude. This score ranges from -1 to 1, where scores near 1 indicate strong alignment between the thermal image and the prompt, while values closer to -1 imply semantic opposition.

Formally, the similarity between an image embedding $\hat{v}_i$ and a text embedding $\hat{u}^c_k$ is defined as:
\begin{equation}
    s(\hat{v}_i, \hat{u}^c_k) = \frac{\hat{v}_i \cdot \hat{u}^c_k}{
    \|\hat{v}_i\| \, \|\hat{u}^c_k\|}
\end{equation}

Hence, each input IR image embedding is compared with the embeddings of prompts representing both present and absent classes. To make classification less sensitive to prompt wording, all prompts in the bank for a class $c \in \{\text{present}, \text{absent}\}$ were averaged to form a centroid embedding:

\begin{equation}
\bar{u}^c = \frac{1}{K} \sum_{k=1}^K \hat{u}^c_k, \quad  
\bar{u}^c \leftarrow \frac{\bar{u}^c}{\|\bar{u}^c\|_2}
\end{equation}

Given a new image $\tilde{x}_i$, the image embedding $\hat{v}_i$ was compared with the centroid embeddings of both classes. The predicted label was assigned to the class with the highest similarity:

\begin{equation}
\hat{y}_i = \arg\max_{c \in \{\text{present}, \text{absent}\}} s(\hat{v}_i, \bar{u}^c)
\end{equation}

The class whose prompts achieve the highest similarity is chosen as the prediction. Hence, the most semantically aligned description will be selected as the label for each image as a class.


\section{Experimental Setup}
\label{sec:experiments}
This section describes the experimental setup used to evaluate the proposed VLM-IRIS framework. The framework combines a VLM with magma preprocessing and centroid-based prompting. To further assess robustness and generalization, alternative settings including different preprocessing strategies, prompting strategies, and dataset conditions were evaluated. In this section, first the dataset and details of variation in object shapes and sizes are introduced. Next, the preprocessing strategies are described. Then, the VLM evaluated in this study is presented. Finally, the prompting strategies applied for zero-shot classification are discussed. 

\begin{figure}[htbp!]
    \centering
    \includegraphics[width=0.8\linewidth, keepaspectratio]{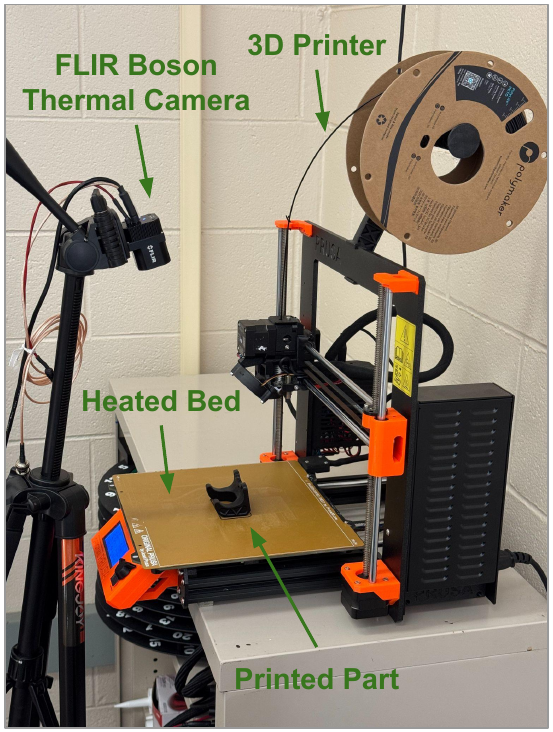}
    \caption{Experimental setup featuring a FLIR Boson thermal camera and
a Prusa MK3S 3D printer on the table.}
    \label{fig:experiment}
\end{figure}

\subsection{Data Collection}

\begin{figure}[htbp!]
    \centering
    \includegraphics[width=\linewidth, keepaspectratio]{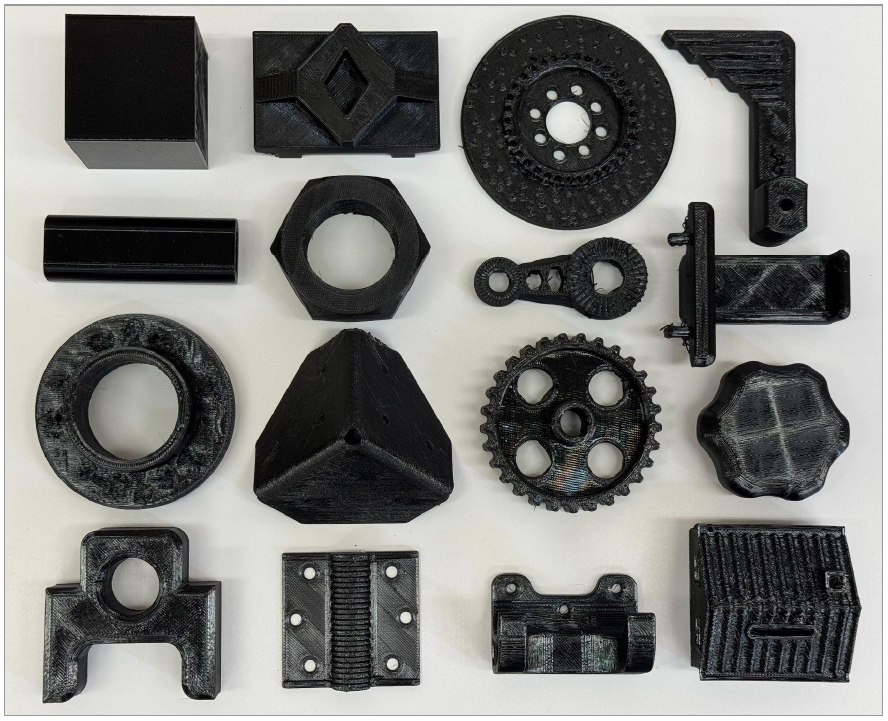}
    \caption{A few 3D printed parts during the experiments. This figure presents the variety of the shapes and geometries of the printed parts.}
    \label{fig:parts}
\end{figure}

The Prusa MK3S 3D printer was used to test the framework as a representative manufacturing platform. A FLIR Boson thermal infrared camera was mounted above the printer's build plate to capture top-down images of the heated bed after printing and provide a consistent view across all experiments. The positioning of the infrared camera can influence the captured thermal patterns, particularly in enclosed build chambers where reflections and occlusions may occur. Since the task focuses on relative thermal contrast for binary object presence detection, no explicit radiometric or geometric calibration was required. Infrared images were captured after print completion to evaluate object presence under controlled thermal conditions. Figure~\ref{fig:experiment} shows the experimental setup that was used for data collection. All test parts were fabricated using PETG filament with the nozzle temperature set to $\sim$230 °C and the bed temperature maintained between 85–110 °C. The collected dataset consists of two classes: object present and object absent. To evaluate the effect of thermal contrast between the surface and the printed part on predictions, data points were captured under a hot build plate ($\sim$85°C) and a room temperature build plate ($\sim$34°C). In total, 200 images were captured, with a balanced distribution across the two categories and temperature conditions. For each printed part, infrared images were captured both immediately after printing on a hot build plate and again after cooling to room temperature. This was done to ensure that the same object geometries were represented under both temperature conditions. To further evaluate generalization, objects with different geometries, shapes, and sizes were printed. The printed parts were intentionally chosen with different shapes and sizes, ranging roughly from 50 to 150 mm in maximum planar extent and 4 to 70 mm in height, to expose the method to a variety of thermal appearances and evaluate how robust the zero-shot object presence detection is across different configurations. Figure~\ref{fig:parts} shows examples of the printed objects used in this study.

\subsection{Preprocessing Variants}

\begin{figure}[htbp!]
    \centering
    \includegraphics[width=\linewidth, keepaspectratio]{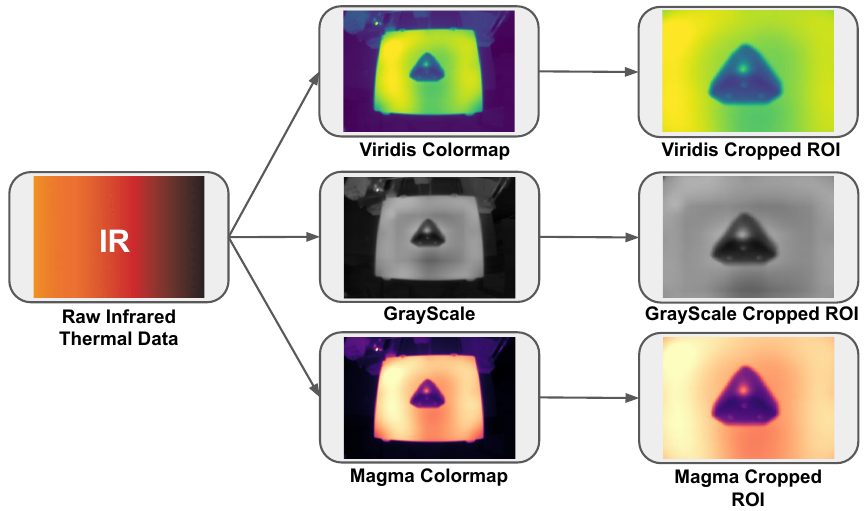}
    \caption{Different preprocessing methods utilized in experiments. The figure presents viridis colormap, grayscale, and magma colormap.}
    \label{fig:preprocess}
\end{figure}

Since VLM are trained on natural RGB images, thermal infrared data must be adapted into a three-channel format before being processed. In this work, three distinct preprocessing strategies are considered: grayscale mapping, magma colormap, and viridis colormap. Figure~\ref{fig:preprocess} visualizes the three different preprocessing methods.

In grayscale mapping, the single thermal intensity channel is normalized and replicated across three channels to preserve the true temperature contrast without adding artificial colors. This method serves as a baseline to test whether VLMs can interpret thermal data that is visually simple yet physically accurate. Magma and viridis colormaps were chosen because they are widely used in scientific visualizations. These colormaps transform intensity values into structured RGB color distributions, where pixel intensity values are mapped to colors. The applied colormaps directly encode the measured temperature values by mapping temperature to color and do not introduce additional information beyond the underlying thermal signal. Such mappings can introduce artificial edges and gradients that may resonate with the color-sensitive filters in pretrained convolutional and transformer encoders. 

While they introduce synthetic colors, they mimic the type of patterns VLMs have seen during pretraining on natural images, which may help the models align thermal data with learned RGB features. Note that testing all three approaches is important because the best option is not obvious. Grayscale stays close to the physical thermal signal, while magma and viridis may better match the color statistics of pretrained models. Hence, this evaluation will determine whether preserving the raw modality or adding color variation leads to stronger zero-shot transfer of VLMs to thermal imagery.

\subsection{Model Configuration}

To evaluate the proposed framework, we used the pretrained CLIP model~\cite{radford2021learning} in a zero-shot configuration, without any retraining or fine-tuning on thermal infrared data. CLIP learns to align images and text through contrastive training for vision–language alignment on hundreds of millions of natural RGB image–caption pairs. Even though it was trained only on RGB images, CLIP has shown strong zero-shot performance across many visual tasks, which makes it a good starting point for testing with thermal infrared imagery. In this study, the ViT-B/32 backbone was selected because it provides a good balance between accuracy and efficiency.

The official PyTorch implementation of the CLIP ViT-B/32 model which employs a Vision Transformer backbone with a 32-pixel patch size was used in all the experiments. The model is pretrained on approximately 400 million image–text pairs collected from the internet. The input resolution is $224\times224$ pixels that are obtained by resizing and center-cropping the preprocessed thermal images. To focus on the region of interest, a center zoom crop is applied with a fraction of 0.50 relative to the smaller image dimension. This fraction ensures that the build plate area is emphasized in the cropped images while it keeps enough surrounding context. Thermal colormaps often introduce strong gradients across the entire build plate which can dominate the embeddings. By zooming in, CLIP sees fewer global gradients from the whole bed and focuses more on the local contrast between the dark object and the bright background. Text embeddings are generated using the same pretrained CLIP text encoder, and all image and text embeddings are normalized to unit length before computing cosine similarity. All experiments were run on CPU, and inference was fast without requiring specialized hardware.

\begin{table*}[t]
\centering
\caption{Zero-shot performance of CLIP (ViT-B/32) across preprocessing, prompting, and datasets. Metrics: Accuracy (Acc), F1, Recall (Rec), Precision (Prec).}
\label{tab:results}
\renewcommand{\arraystretch}{2} 
\setlength{\tabcolsep}{11pt}       
\small                             
\begin{tabular}{l l *{8}{c}}
\toprule
& & \multicolumn{4}{c}{\textbf{Hot}} & \multicolumn{4}{c}{\textbf{Room Temperature}} \\
\cmidrule(lr){3-6}\cmidrule(lr){7-10}
\textbf{Preproc.} & \textbf{Prompting} & Acc & F1 & Rec & Prec & Acc & F1 & Rec & Prec \\
\midrule
\multirow{2}{*}{Grayscale} & Single   & 70.00 & 65.12 & 56.00 & 77.78 & 97.00 & 96.91 & 94.00 & 100.00 \\
                           & Centroid & 83.00 & 79.52 & 66.00 & 100.00 & 99.00& 98.99 & 98.00 & 100.00 \\
\midrule
\multirow{2}{*}{Magma}     & Single   & 83.00 & 79.52 & 66.00 & 100.00 & 92.00 & 92.16 & 94.00 & 90.38 \\
                           & Centroid & 92.00 & 91.30 & 84.00 & 100.00 & \textbf{100.00} & \textbf{100.00} & \textbf{100.00} & \textbf{100.00} \\
\midrule
\multirow{2}{*}{Viridis}   & Single   & 78.00 & 71.79 & 56.00 & 100.00 & 93.00 & 92.47 & 86.00 & 100.00 \\
                           & Centroid & 88.00 & 89.09 & 98.00 & 81.67 & 95.00 & 95.24 & 100.00 & 90.91 \\
\bottomrule
\end{tabular}
\end{table*}

\begin{figure}[htbp!]
    \centering
    \includegraphics[width=0.98\linewidth, keepaspectratio]{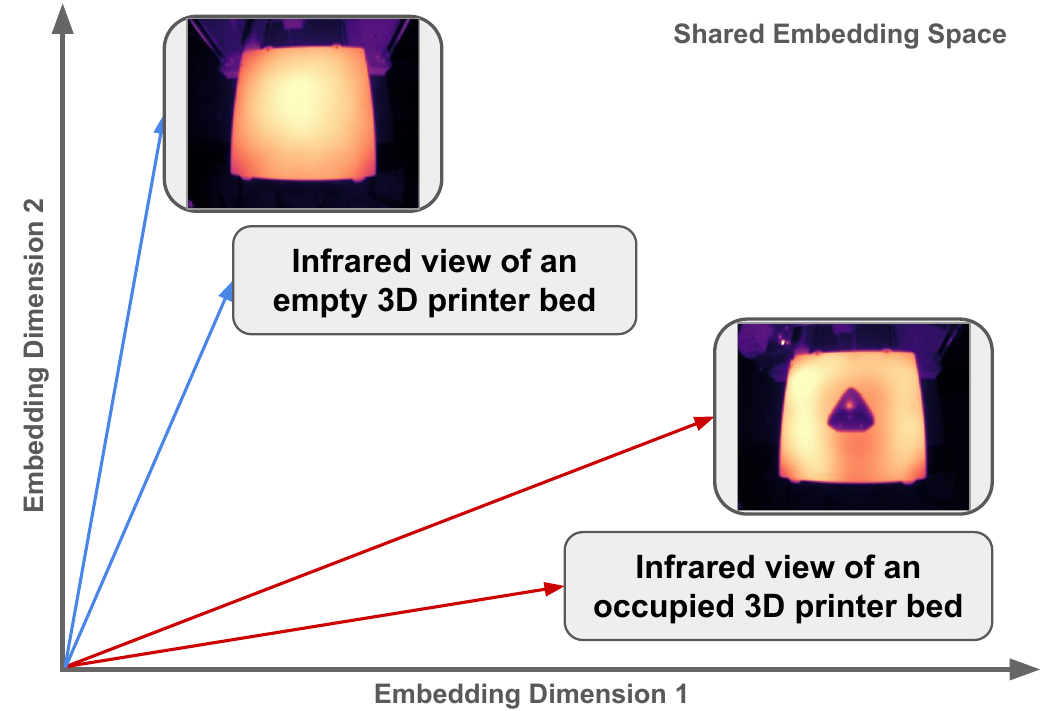}
    \caption{Example of image-text pairs in shared embedding space. Infrared images of an empty and an occupied 3D printer are projected alongside their textual prompts, showing how the alignment of vectors reflects semantic similarity.}
    \label{fig:embeding}
\end{figure}
\subsection{Prompting Strategies}

In VLMs, images and text are mapped into a shared embedding space where semantic similarity can be directly measured. Each infrared image is encoded into a feature vector and compared against feature vectors generated from text prompts that describe possible scene interpretations. The prompt whose embedding is most similar to the image embedding determines the predicted class. Figure~\ref{fig:embeding} illustrates this process in a simplified two-dimensional embedding space. VLMs work by matching what is seen in an image with what is described in text. In this context, prompts written by humans are used to clearly describe the task using visual cues that people naturally notice such as contrast, color, and whether a surface is occupied by an object. More descriptive and task-relevant prompts help guide the model toward relevant visual patterns. Prompt design is important since it defines how the model links thermal image features with semantic meaning. Because different word choices can influence the outcome, this work employs a prompt bank, where embeddings from multiple prompts are averaged to form a more stable representation for each class and reduce sensitivity to individual wording. Two classes were defined: present, where an object is on the build plate, and absent, where the build plate is empty. Several sentences that describe the same scene in different ways were composed to form the prompt bank for each class. In this bank, prompts for the present class emphasize a clear dark object resting on the heated surface, while prompts for the absent class focus on the surface being empty. For instance, prompts such as “an infrared image showing a bright surface with a solid shape on it” or “a thermal infrared image of a bright surface occupied by a solid object” were used to describe the present class. For the absent class, prompts like “infrared photo of a flat bright surface that is completely empty” or “a thermal image of a bright platform with no blocks or parts” were used. To keep the experiments robust to different phrasings, the same prompt banks were reused across preprocessing methods with only minor wording changes. For magma preprocessing, descriptions were adjusted to refer to an orange surface instead of bright, while for viridis preprocessing the prompts referred to a green–yellow surface.

To test the impact of prompt banking, a simpler single-prompt inference strategy was also evaluated. In this setup, each class was represented by one prompt selected from its corresponding prompt bank. For the present class, the prompt “infrared view of a bright flat surface where an object sits on top” was used, and for the absent class, the prompt “thermal image of a bright background with no solid objects on it” was chosen.
To identify the most representative prompt for each class, the cosine similarity scores between all images and every prompt in the bank were calculated. The prompt that achieved the highest average similarity across images was selected as the single-prompt sentence. Classification was then based on the similarity between the image embedding and these two fixed prompt embeddings. In contrast, the centroid inference strategy uses multiple prompts for each class. All prompt embeddings are averaged into a single centroid representation, and classification is based on the similarity between the image embedding and these centroids. This acts as a form of prompt ensembling to reduce the effect of prompt-to-prompt variability and providing a more stable and robust representation of each category.





\section{Results}
\label{sec:results}

Table~\ref{tab:results} summarizes the quantitative results. Overall, the results confirm that VLMs can interpret thermal data when properly adapted through preprocessing and prompt design. Among all tested configurations, the best performance was achieved using magma preprocessing with centroid prompting on the room-temperature dataset, reaching 100\% accuracy, precision, and recall. However, the type of preprocessing, the prompting strategy, and the thermal state of the bed had a clear impact on how CLIP interpreted the scenes that are discussed below. 

The temperature of the printer bed had a strong effect on the classification results. All preprocessing methods performed better at room temperature than when the bed was hot. Under hot-bed conditions, both the bed and the printed part are very warm, and the temperature difference between them becomes small and gradual. This makes the boundaries appear softer to CLIP, even though they are still visible to the human eye. Such thermal reflections and glow patterns are visually different from the clean sharp contrasts between objects and backgrounds seen in its training data, which likely confused the model. Prompts like “a block on a bright surface” assume clean patterns and not extreme thermal gradients or sensor noise seen in infrared imagery. This makes object boundaries even less distinct in grayscale or low-color images and challenges the VLMs ability to separate the two regions.

\begin{figure*}[htb!]
    \centering
    \includegraphics[width=\linewidth, keepaspectratio]{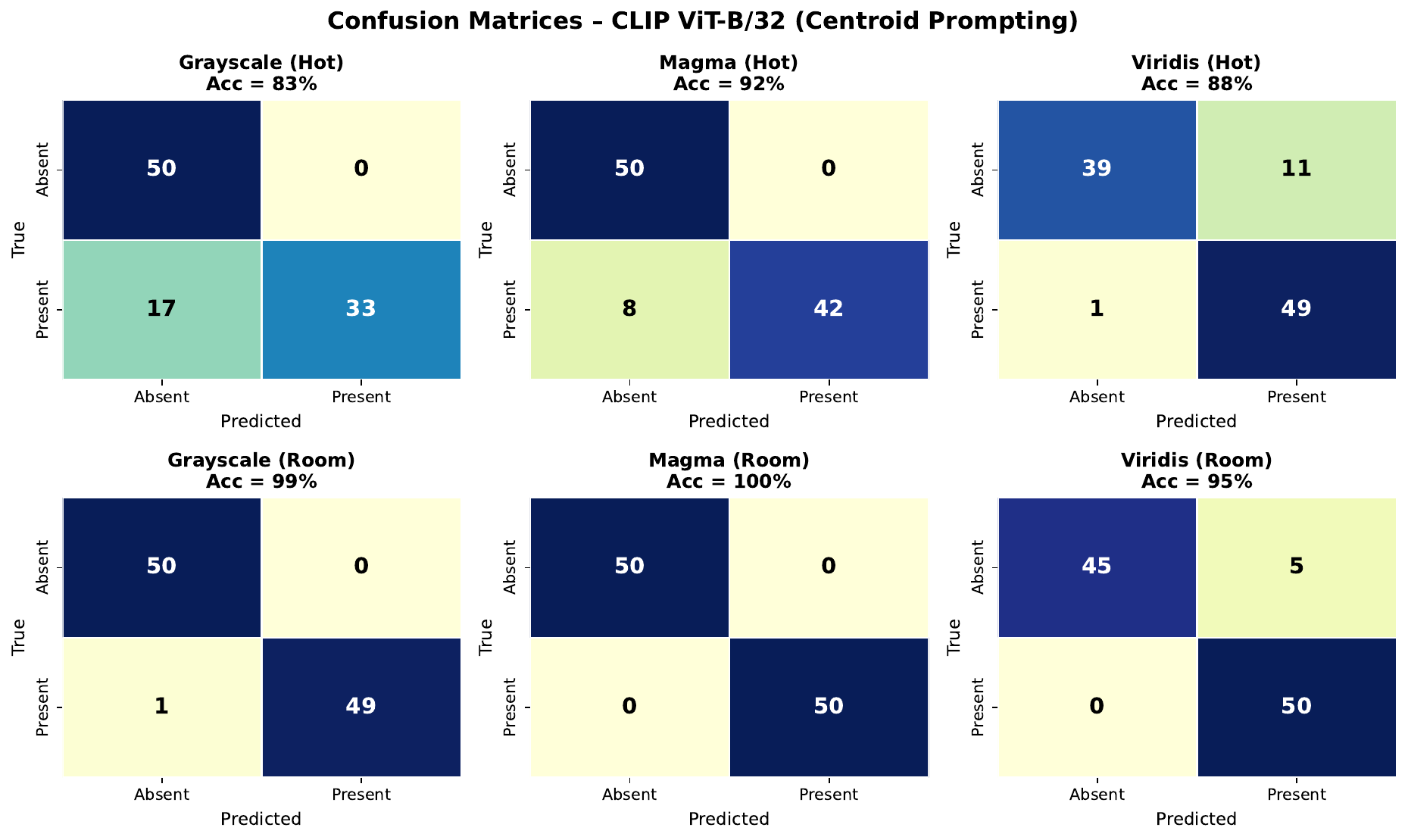}
    \caption{Confusion matrices for CLIP ViT-B/32 under centroid prompting across preprocessing methods and temperature conditions. The top row shows results under hot-bed (~85 °C) conditions, and the bottom row under room temperature (~34 °C).}
    \label{fig:conf}
\end{figure*}

The grayscale preprocessing achieved 83\% accuracy on the hot-bed dataset, with perfect precision (100\%) but lower recall (66\%). This means that whenever the model predicted an object was present, it was correct, but it missed some true “present” cases. The confusion matrix in Figure \ref{fig:conf} shows 17 such misses where the model labeled real objects as absent. In contrast, colormaps such as magma and viridis performed better by providing artificial color gradients that give the model more familiar cues that help it separate the printed part from the bed. In other words, the added colors make the infrared images look more like what CLIP expects, which leads to improving recognition. The magma colormap achieved 92\% accuracy, where all empty surfaces were correctly recognized as absent, and only 8 “present” samples were missed. The viridis preprocessing also achieved strong results, with 88\% accuracy, while the model correctly identified nearly all “present” cases (missing only one) but incorrectly labeled 11 empty-bed images as “present.” This suggests that the green–yellow gradient in viridis can introduce textures that CLIP misinterprets as object edges or surface irregularities, especially when the bed emits uneven heat. Overall, in hot conditions, the model tends to confuse reflections or hot zones with object presence, and color mappings like magma help reduce this effect.

When the bed cools to room temperature, the thermal gradients become more stable and they produce embeddings that better fit CLIP’s pretraining distribution that leads to higher cosine similarity scores and more accurate recognition. This happens because the scene looks more similar to a normal RGB photo, and CLIP’s learned RGB filters respond more effectively to it and can match the image to the prompts more accurately. These smoother gradients help the image become less noisy and the model can pay more attention to the object’s shape and edges instead of being distracted by bright or uneven heat areas. Figure \ref{fig:conf} shows the prediction results using a confusion matrix.

Under room temperature conditions, the grayscale preprocessing achieved almost perfect performance with 99\% accuracy and only one “present” image that was incorrectly labeled as “absent.” This shows that when the thermal environment is stable, CLIP can clearly separate the printed part from the bed even without any added color information. The grayscale mapping keeps the true thermal information intact and avoids artifacts that can come from artificial coloring that makes it a strong and reliable choice when the temperature contrast is moderate and consistent. The magma colormap achieved perfect performance across all metrics (100\% accuracy, precision, and recall). When the thermal range becomes narrower at room temperature, magma’s smooth gradient is able to represent even small differences across the surface through color cues that CLIP’s filters can use to detect the edges and textures of the object. This suggests that magma’s color scale fits well with the kind of color structure CLIP was trained on that makes it the most robust and balanced method across both temperature conditions. The viridis colormap also performed strongly by reaching 95\% accuracy. However, five empty-bed images were incorrectly classified as “present.” The lower temperature range allows viridis to use a more balanced color gradient that reduces the false positives by making the bed’s texture smoother. However, compared to magma and grayscale, viridis can still exaggerate small intensity differences into artificial patterns, which explains the occasional false detections. Overall, viridis performs well but is slightly more sensitive to noise in uniform regions compared to magma.
\section{Conclusions}
Across all preprocessing methods, the centroid prompting strategy performed better than using a single prompt. By averaging multiple text descriptions into one combined representation, the model becomes less sensitive to how each sentence is worded and produces more stable results. Overall, using the centroid prompts gave more reliable predictions, especially for thermal images where small changes in phrasing can strongly affect how the model interprets the scene.


\label{sec:conclusions}

This work demonstrates a notable advancement in the field of scene description for manufacturing by presenting VLM-IRIS, a zero-shot framework that adapts VLMs for use with thermal infrared imagery. This work developed and validated a framework that converts infrared data into RGB-compatible images and pairs them with descriptive text prompts, and does object presence detection without any retraining or labeled datasets. VLM-IRIS operates fully in a zero-shot manner, which makes it easy to maintain and reliable over time, and allows it to keep working well even if the equipment or environment changes. The framework is not tied to a specific material extrusion printer or filament material since it relies on relative thermal contrast and semantic alignment. When applied to different printers or materials, factors including emissivity, surface finish, and operating temperature ranges may affect how the infrared images appear and how prompts should be written. However, the core zero-shot pipeline would remain unchanged, and therefore adapting the method mainly involves updating the prompt descriptions rather than retraining the model. This creates new opportunities for using such models in manufacturing, where thermal cameras are already common for process monitoring, fault detection, and inspection.

\vfill\pagebreak
The experiments on a 3D printer testbed demonstrated that CLIP (ViT-B/32) can effectively interpret thermal scenes when supported by suitable preprocessing and prompt design. The experiments confirmed the importance of the preprocessing method and highlighted the impact of prompt engineering on model predictions. Among the tested approaches, the magma colormap with centroid-based prompting achieved the best results, reaching 100\% accuracy under room-temperature conditions. Grayscale mapping provides the most physically accurate and interpretable representation but can miss objects when thermal contrast is weak. Magma delivers the most consistent performance across conditions by introducing structured color gradients that resemble RGB data CLIP was trained on. Viridis also performs well overall but tends to generate false positives in high-temperature scenes due to reflective heat patterns. In terms of the temperature of the setting, moderate contrast and structured color mappings help CLIP form more stable embeddings, while extreme heat gradients add visual noise that reduces accuracy. Future work will explore how these findings transfer to other VLMs, since differences in architecture, training data, and embedding behavior may influence sensitivity to preprocessing choices and prompt design.

The object presence detection scenario considered in this study is not intended to represent the full complexity of industrial perception tasks. Rather, it was selected to isolate the challenge of adapting thermal infrared imagery to VLMs in a zero-shot setting. The selected scenario allows us to examine the effects of preprocessing and prompt design without introducing additional complexity from multi-class recognition or geometric reasoning. At the same time, object presence detection remains a practically useful capability for downstream automation in manufacturing environments. As the results confirm that VLMs which are originally designed for natural RGB imagery can be effectively adapted to thermal infrared domains, the framework will be extended to more complex inspection tasks, such as defect detection in additive manufacturing. In future extensions, different defect levels, such as small, medium, or large, can be described with text prompts, and the model can identify the most likely category by comparing the infrared image with these prompt descriptions, without retraining the model.

The proposed methodology has the potential to extend beyond material extrusion processes to a range of additive manufacturing modalities where thermal sensing is commonly employed. The robustness of the approach arises from its reliance on relative thermal patterns and semantic reasoning through VLMs, rather than process-specific training or handcrafted features. However, different additive manufacturing processes can show different thermal behavior and surface characteristics, which may affect infrared appearance. Adapting the method to these settings may require adjusting the preprocessing steps or refining the prompts, which is an important direction for future work. Center cropping in the experiments assumes a fixed camera view and a known build-plate region, which was used here to focus on infrared modality adaptation and zero-shot inference without adding localization complexity. In more general settings, the same framework could be extended by evaluating multiple regions or integrating localization mechanisms, which is left for future work. Moreover, future work will explore fine-tuning these models on thermal datasets\cite{ghiasvand2025few,kermani2025systematic}. This fine-tuning will help the model better learn the characteristics of infrared imagery in industrial environments, which will improve both accuracy and robustness of the framework. In addition, incorporating controlled positive noise into the transformer architecture may help reduce model entropy and simplify classification~\cite{esfandiari2025multi}. Finally, explainable-AI techniques can provide insight into what visual cues CLIP relies on when embedding the image and text that helps guide further optimization of the framework.




\nocite{*}

\bibliographystyle{asmeconf}  
\bibliography{asmeconf-sample}

\end{document}